\def\year{2021}\relax
\documentclass[letterpaper]{article} 
\usepackage{aaai21} 
\usepackage{times} 
\usepackage{helvet} 
\usepackage{courier} 
\usepackage[hyphens]{url} 
\usepackage{graphicx} 
\usepackage{amsmath}
\usepackage{esvect}
\usepackage{array}
\usepackage{pgf-pie}
\usepackage{tabulary}
\newcolumntype{K}[1]{>{\centering\arraybackslash}p{#1}}
\usepackage{natbib} 
\usepackage{caption} 
\usepackage[switch]{lineno}
\title{Acronym Identification and Disambiguation Shared Tasks\\for Scientific Document Understanding}

\author{
 Amir Pouran Ben Veyseh\textsuperscript{\rm 1}, Franck Dernoncourt\textsuperscript{\rm 2}, Thien Huu Nguyen\textsuperscript{\rm 1}, \\ Walter Chang\textsuperscript{\rm 2}, and Leo Anthony Celi\textsuperscript{\rm 3,4} \\
\textsuperscript{\rm 1}{\normalfont University of Oregon, Eugene, OR, USA}\\
\textsuperscript{\rm 2}{\normalfont Adobe Research, San Jose, CA, USA}\\
\textsuperscript{\rm 3}{\normalfont Harvard University, Cambridge, MA, USA}\\
\textsuperscript{\rm 4}{\normalfont Massachusetts Institute of Technology, Cambridge, MA, USA}\\

{\tt \{thien, apouranb\}@cs.uoregon.edu} \\
{\tt \{franck.dernoncourt, wachang\}@adobe.com} \\
{\tt lceli@bidmc.harvard.edu}
}

\begin{document}
\maketitle

\begin{abstract}
Acronyms are the short forms of longer phrases and they are frequently used in writing, especially scholarly writing, to save space and facilitate the communication of information. As such, every text understanding tool should be capable of recognizing acronyms in text (i.e., acronym identification) and also finding their correct meaning (i.e., acronym disambiguation). As most of the prior works on these tasks are restricted to the biomedical domain and use unsupervised methods or models trained on limited datasets, they fail to perform well for scientific document understanding. To push forward research in this direction, we have organized two shared task for acronym identification and acronym disambiguation in scientific documents, named AI@SDU and AD@SDU, respectively. The two shared tasks have attracted 52 and 43 participants, respectively. While the submitted systems make substantial improvements compared to the existing baselines, there are still far from the human-level performance. This paper reviews the two shared tasks and the prominent participating systems for each of them.
\end{abstract}

\section{Introduction}
One of the common practices in writing to save space and make the flow of information smoother is to avoid repetition of long phrases which might waste space and the reader's time. To this end, acronyms that are the shortened form of a long-phrase are often used in various types of writing, especially in scientific documents. However, this prevalence might introduce more challenges for text understanding tools. More specifically, as the acronyms might not be defined in dictionaries, especially locally-defined acronyms whose long-form is only provided in the document that introduces them, a text processing model should be able to identify the acronyms and their long forms in the text (i.e., acronym identification). For instance, in the sentence ``\textit{The main key
performance indicator, herein referred to as KPI, is the E2E throughput}", the text processing system must recognize \textit{KPI} and \textit{E2E} as acronyms and the phrase \textit{key performance indicator} as the long-form. Another issue related to the acronym that text understanding tools encounter is that the correct meaning (i.e., long-form) of the acronym might not be provided in the document itself (e.g., the acronym \textit{E2E} in the running example). In these cases, the correct meaning can be obtained by looking up the meaning in an acronym dictionary. However, as different long forms could share the same acronym (e.g., two long forms \textit{Cable News Network} and \textit{Convolution Neural Network} share the acronym \textit{CNN}), this meaning look-up is not straightforward and the system must disambiguate the acronym (i.e., acronym disambiguation). Both AI and AD models could be used in downstream applications including definition extraction \cite{veyseh2020joint,spala-etal-2020-semeval,spala-etal-2019-deft,espinosa-anke-schockaert-2018-syntactically,jin-etal-2013-mining}, various information extraction tasks \cite{liu2019gcdt,pouran-ben-veyseh-etal-2019-graph} and  question answering \cite{ackermann2020resolution,veyseh2016cross}.  

Due to the importance of the two aforementioned tasks, i.e. acronym identification (AI) and acronym disambiguation (AD), there is a wealth of prior work on AI and AD \cite{park2001hybrid,schwartz2002simple,nadeau2005supervised,kuo2009bioadi,taneva2013mining,kirchhoff2016unsupervised,li2018guess,ciosici2019unsupervised,jin2019deep}. However, there are two major limitations in the existing systems. First, for AD tasks, the existing models are mainly limited to the biomedical domain, ignoring the challenges in other domains. Second, for the AI task, the existing models employ either unsupervised methods or models trained using a limited manually annotated AI dataset. The unsupervised methods or small size of the AI dataset results in errors for acronym identification which could be also propagated for acronym disambiguation task.

\begin{table*}[ht]
\centering
\resizebox{.98\textwidth}{!}{
\begin{tabular}{l|c|c|c|c|c|c}
Dataset & Size & \# Unique Acronyms & \# Unique Meaning & \# Documents & Publicly Available & Domain \\ \hline
LSAEF \cite{liu2011learning} & 6,185 & 255 & 1,372 & N/A & No & Wikipedia \\
AESM \cite{nautial2014finding} & 355 & N/A & N/A & N/A & No & Wikipedia \\
MHIR \cite{harris2019my} & N/A & N/A & N/A & 50 & No & Scientific Papers \\
MHIR \cite{harris2019my} & N/A & N/A & N/A & 50 & No & Patent \\
MHIR \cite{harris2019my} & N/A & N/A & N/A & 50 & No & News \\
SciAI (ours) & 17,506 & 7,964 & 9,775 & 6,786 & yes & Scientific Papers
\end{tabular}
}
\caption{Comparison of non-medical manually annotated acronym identification datasets. Note that size refers to the number of sentences in the dataset.}
\label{tab:compare-AI}
\end{table*}

\begin{table*}[ht]
\centering
\resizebox{.98\textwidth}{!}{
\begin{tabular}{l|c|c|c}
Dataset & Size & Annotation & Avg. Number of Samples per Long Form \\ \hline
Science WISE \cite{prokofyev2013ontology} & 5,217 & Disambiguation manually annotated & N/A \\
NOA \cite{charbonnier2018using} & 19,954 & No manual annotation & 4 \\
SciAD (ours) & 62,441 & Acronym identification manually annotated & 22
\end{tabular}
}
\caption{Comparison of scientific acronym disambiguation (AD) datasets. Note that size refers to the number of sentences in the dataset.}
\label{tab:compare-AD}
\end{table*}

To address the above issues in the prior works, we recently released the largest manually annotated acronym identification dataset for scientific documents (viz., SciAI) \cite{veyseh-et-al-2020-what}. This dataset consists of 17,506 sentences from 6,786 English papers published in arXiv. The annotation of each sentence involves the acronyms and long forms mentioned in the sentence. Also, using this manually annotated AI dataset, we also created a dictionary of 732 acronyms with multiple corresponding long forms (i.e., ambiguous acronyms) which is the largest available acronym dictionary for scientific documents. Moreover, using the prepared dictionary and 2,031,592 sentences extracted from arXiv papers, we created a dataset for the acronym disambiguation task (viz., SciAD) \cite{veyseh-et-al-2020-what}. This dataset consists of 62,441 sentences, which is larger than the prior AD dataset for the scientific domain.

Using the two datasets SciAI and SciAD, we organize two shared tasks for acronym identification and acronym disambiguation for scientific document understanding (i.e., AI@SDU and AD@SDU, respectively). The AI@SDU shared task has attracted 52 participant teams with 19 submissions during the evaluation phase. The AD@SDU has also attracted 43 participant teams with 10 submissions during the evaluation phase. The participant teams made considerable progress on both shared task compared to the provided baselines. However, the top-performing models, (viz., \textit{AT-BERT-E} for AI@SDU with 93.3\% F1 score and \textit{DeepBlueAI} for AD@SDU with 94.0\% F1 score), underperforms human (with 96.0\% and 96.1\% F1 score for AI@SDU and AD@SDU shared task, respectively), leaving room for future research. In this paper, we review the dataset creation process, the details of the shared task, and the prominent submitted systems.

\section{Dataset \& Task Description}

\subsection{Acronym Identification}
The acronym identification (AI) task aims to recognize all acronym and long forms mentioned in a sentence. Formally, given the sentence $S=[w_1,w_2,\ldots,w_N]$ the goal is to predict the sequence $L=[l_1,l_2,\ldots,l_N]$ where $l_i\in\{B_a,I_a,B_l,I_l,O\}$. Note that $B_a$ and $I_a$ indicate the beginning and inside an acronym, respectively, while $B_l$ and $I_l$ show beginning and inside a long form, respectively, and $O$ is the label for other words. 

As mentioned in the introduction, the existing AI datasets are either created using some unsupervised methods (e.g., by character matching the acronym with their surrounding words in the text) or they are small-sized thus inappropriate for data-hungry deep learning models. To address these limitations we aim to create the largest acronym identification dataset which is manually labeled. To this end, we first collect 6,786 English papers from arXiv. This collection contains 2,031,592 sentences. As all of these sentences might not contain the acronym and their long forms, we first filter out the sentences without any candidate acronym and long-form. To identify the candidate acronyms, we use the rule that the word $w_t$ is a candidate acronym if half of its characters are upper-cased. To identify the candidate long forms, we employ the rule that the subsequent words $[w_j,w_j+1,\ldots,w_{j+k}]$ are a candidate long-form if the concatenation of their first one, two, or three characters can form a candidate acronym, i.e., $w_t$, in the sentence. After filtering sentences without any candidate acronym and long-form, 17,506 sentences are selected that are annotated by three annotators from Amazon Mechanical Turk (MTurk). More specifically, MTurk workers annotated the acronyms, long forms, and the mapping between identified acronyms and long forms. In case of disagreements, if two out of three workers agree on an annotation, we use majority voting to decide the correct annotation. Otherwise, a fourth annotator is hired to resolve the conflict. The inter-annotator agreement (IAA) using Krippendorff’s alpha \cite{krippendorff2011computing} with the MASI distance metric \cite{passonneau2006measuring} for short-forms (i.e., acronyms) is 0.80 and for long-forms (i.e., phrases) is 0.86. This dataset is called SciAI. A comparison of the SciAI dataset with other existing manually annotated AI datasets is provided in Table \ref{tab:compare-AI}. 

\subsection{Acronym Disambiguation}
The goal of acronym disambiguation (AD) task is to find the correct meaning of a given acronym in a sentence. More specifically, given the sentences $S=[w_1,w_2,\ldots,w_N]$ and the index $t$ where $w_t$ is an acronym with multiple long forms $L=\{l_1,l_2,\ldots,l_m\}$ the goal is to predict the long form $l_i$ form $L$ as the correct meaning of $w_t$. 

As discussed earlier, one of the issues with the existing AD datasets is that they mainly focus on the biomedical domain, ignoring the challenges in other domains. This domain shift is important as some of the existing models for biomedical AD exert domain-specific resources (e.g., BioBERT) which might not be suitable for other domains. Another issue of the existing AD datasets, especially the ones proposed for a scientific domain, is that they are based on unsupervised AI datasets. That is, acronyms and long forms in a corpus are identified using some rules and the resulting AI dataset is employed to find acronyms with multiple long forms to create the AD dataset. This unsupervised method to create an AD dataset could introduce noises and miss some challenging cases. To address these limitations, we created a new AD dataset using the manually labeled SciAI dataset. More specifically, first using the mappings between annotated acronyms and long forms in SciAI, we create a dictionary of acronyms that have multiple long forms (i.e., ambiguous acronyms). This dictionary contains 732 acronyms with an average of 3.1 meaning (i.e., long-form) per acronym. Afterward, to create samples for the AD dataset, we look up all sentences in the collected corpus in which one of the ambiguous acronyms is locally defined (i.e., its long-form is provided in the same sentence). Next, in the documents hosting these sentences, we automatically annotate every occurrence of the acronym with its locally defined long-form. Using this process a dataset consisting of 62,441 sentences is created. We call this dataset SciAD. A comparison of the SciAD dataset with other existing scientific AD dataset is provided in Table \ref{tab:compare-AD}

\section{Participating Systems \& Results}
\subsection{Acronym Identification}
For the AI task, we provide a rule-based baseline. In particular, inspired by \cite{schwartz2002simple}, the baseline identifies the acronyms and their long-forms if they match one of the patterns of \textit{long form (acronym)} or \textit{acronym (long form)}. More specifically, if there is a word with more than 60\% upper-cased characters which is inside parentheses or right before parentheses, it is predicted as an acronym. Afterward, we assess the words before or after the acronym (depending on which pattern the predicted acronym belongs to) that fall into the pre-defined window of size $\text{min}(|A|+5,2*|A|)$, where $|A|$ is the number of characters in the acronym. In particular, if there is a sequence of characters in these words which can form the upper-cased characters in the acronym then the words after or before the acronym are selected as its meaning (i.e., long-form). Moreover, as SciAI dataset annotates acronyms even if they do not have any locally defined long-form, we extend the rule for identifying the acronyms by relaxing the requirement of being inside or before the parentheses.

\begin{table}[]
\centering
\resizebox{.49\textwidth}{!}{
\begin{tabular}{l|ccc}
\textbf{Team Name} & Precision & Recall & F1 \\ \hline
GCDH & 86.50 & 85.57 & 86.03 \\
Aadarshsingh & 88.26 & 89.08 & 88.67 \\
TAG-CIC & 89.70 & 88.16 & 88.92 \\
Spark & 89.91 & 90.49 & 90.20 \\
Dumb-AI & 89.72 & 90.94 & 90.33 \\
Napsternxg & 90.15 & 91.15 & 90.65 \\
Pikaqiu & 91.02 & 90.51 & 90.76 \\
SciDr \cite{Aadarsh2020SciDr} & 90.98 & 90.83 & 90.90 \\
Aliou & 90.78 & 91.12 & 90.95 \\
AliBaba2020 & 90.30 & 92.87 & 91.57 \\
RK & 89.93 & 93.88 & 91.86 \\
DeepBlueAI & 92.01 & 91.84 & 91.92 \\
EELM-SLP \cite{Kubal2020Effective} & 89.70 & 94.59 & 92.08 \\
Lufiedby & 92.64 & 91.74 & 92.19 \\
Primer \cite{Nicholas2020Primer} & 91.73 & 93.49 & 92.60 \\
HowToSay & 91.93 & 93.70 & 92.81 \\
N\&E \cite{Feng2020Systems} & 93.49 & 92.74 & 93.11 \\
AT-BERT-E \cite{Danqing2020ATBERT} & 92.20 & 94.43 & \textbf{93.30} \\ \hline
Baseline (Rule-based) & 91.31 & 77.93 & 84.09 \\ \hline
Human Performance & 97.70 & 94.56 & 96.09
\end{tabular}
}
\caption{Performance of the participating systems in Acronym Identification task}
\label{tab:ai}
\end{table}

In the AI@SDU task, 54 teams participated and 18 of them submitted their system results in the evaluation phase. In total, all teams submitted 254 submissions for different versions of their models. The submitted systems employ various methods including:
(1) \textbf{Rule-based Methods}: Similar to our baseline, some participants exploited manually designed rules which could have high precision, but low recall \cite{Willie2020AINLM, Feng2020Systems} (2) \textbf{Feature-based Models}: These models extract various features from the texts to be used by a statistical model to predict the acronyms and long forms \cite{Feng2020Systems} (3) \textbf{Transformer-based models}: In these systems, the sentence is encoded with a pre-trained transformer-based language model and the labels are predicted using the obtained word embeddings \cite{Kubal2020Effective, Feng2020Systems, Nicholas2020Primer}. Some of these models may also leverage adversarial training to make the model more robust to the noises \cite{Danqing2020ATBERT} or they might employ an ensemble model \cite{Aadarsh2020SciDr}. Among all submitted models, the method proposed by \cite{Danqing2020ATBERT}, i.e., AT-BERT-E, achieves the highest performance. This model employs an adversarial training approach to increase the model robustness toward the noise. More specifically, they augment the training data with adversarial perturbed samples and fine-tune a BERT model followed by a feed-forward neural net on this task. For the adversarial perturbation, they leverage a gradient-based approach in which the sample representations are altered in the direction that the gradient of the loss function rises.

We evaluate the systems based on macro-averaged precision, recall, and F1 score of the acronym and long-form prediction. The results are shown in Table \ref{tab:ai}. This table shows that the participants have made considerable improvement over the provided baseline. However, there is still a gap between the performance of the task winner (i.e., AT-BERT-E \cite{Danqing2020ATBERT}) and human-level performance, suggesting more improvement is required. 

\subsection{Acronym Disambiguation}
For AD task, we propose to employ the frequency of the acronym long forms to disambiguate them. More specifically, for the acronym $a$ with the long forms $L=[l_1,l_2,\ldots,l_m]$, we compute the number of occurrence of each of its long forms in the training data, i.e., $F=[f_1,f_2,\ldots,f_m]$ where $f_i = |\mathcal{A}^a_i|$ and $\mathcal{A}^a_i$ is the set of sentences in the training data with the acronym $a$ and the long form $l_i$. In inference time, the acronym $a$ is expanded to its long form with the highest frequency, i.e., $i^*=\text{arg max}_i f_i$.

\begin{table}[]
\centering
\resizebox{.49\textwidth}{!}{
\begin{tabular}{l|ccc}
\textbf{Team Name} & Precision & Recall & F1 \\ \hline
UC3M \cite{Areej2020Participation} & 92.15 & 77.97 & 84.37 \\
AccAcE \cite{Joao2020Acronym} & 93.57 & 83.77 & 88.40 \\
GCDH & 94.88 & 87.03 & 90.79 \\
Spark & 94.87 & 87.23 & 90.89 \\
AI-NLM \cite{Willie2020AINLM} & 90.73 & 91.96 & 91.34 \\
Primer \cite{Nicholas2020Primer} & 94.72 & 88.64 & 91.58 \\
Sansansanye & 95.18 & 88.93 & 91.95 \\
Zhuyeu & 95.48 & 89.07 & 92.16 \\
Dumb AI & 95.95 & 89.59 & 92.66 \\
SciDr \cite{Aadarsh2020SciDr} & 96.52 & 90.09 & 93.19 \\
hdBERT \cite{Qiwei2020Leveraging} & 96.94 & 90.73 & 93.73 \\
DeepBlueAI \cite{Chunguang2020BERTbased} & 96.95 & 91.32 & \textbf{94.05} \\ \hline
Baseline (Freq.) & 89.00 & 46.36 & 60.97 \\ \hline
Human Performance & 97.82 & 94.45 & 96.10
\end{tabular}
}
\caption{Performance of the participating systems in Acronym Disambiguation task}
\label{tab:ad}
\end{table}

The AD@SDU task attracted 44 participants, 12 submissions at the evaluation phase, and 187 total submissions for different versions of the participating systems. This task has been approached with a variety of methods, including (1) \textbf{Feature-based models}: Some systems extract features from the input sentence (e.g., word stems, part-of-speech tags, or special characters in the acronym). Next, a statistical model, such as Support Vector Machine, Naive Bayes, and K-nearest neighbors, is employed to predict the correct long form of the acronym \cite{Areej2020Participation, Joao2020Acronym}; (2) \textbf{Neural Networks}: A few of the participating systems employ deep architectures, e.g., convolution neural networks (CNN) or long short-term memory (LSTM) \cite{Willie2020AINLM}; (3) \textbf{Transformer-based Models}: The majority of the participants resort to transformer-based language models, e.g., BERT, SciBERT or RoBERTa, to encode the input sentence. However, they differ in how they leverage the outputs of these language models for prediction and also how they formulate the task. Whereas most of the existing works formulate the task as a classification problem \cite{Chunguang2020BERTbased, Qiwei2020Leveraging}, authors in \cite{Nicholas2020Primer} use an information retrieval approach. More specifically, the cosine similarity between the embeddings of the candidates and the input is employed to compute the score of each candidate and then to rank them based on their scores. Moreover, authors in \cite{Aadarsh2020SciDr} model this task as a span prediction  problem. Specifically, the concatenation of the different candidate long forms with the acronym and the input sentence is encoded by a transformer-based language model. Afterward, a sequence labeling component predicts the sub-sequence with the highest probability of being the correct long form. Among all submitted systems, the DeepBlueAI model proposed by \cite{Chunguang2020BERTbased} obtained the highest performance for acronym disambiguation on SciAD test set. This model formulate this task as a binary classification problem in which each candidate long-form is assigned a score by a binary classifier and the candidate with the highest scores is selected as the final model prediction. For the classifier, authors employ a pre-trained BERT model that takes the input in the form of $L_i\text{ }[SEP]\text{ }w_1,w_2,\ldots,start,w_a,end,\ldots,w_n$, where $L_i$ is the long-form candiate, $w_i$ is the words of the input sentence, $w_a$ is the ambiguous acronym in the input sentence, and $start$ and $end$ are two special tokens to provide the position of the acronym to the model. 

We evaluate the systems using their macro-averaged precision, recall, and F1 score for predicting the correct long form. The results are shown in Table \ref{tab:ad}. Again, this table shows that the participating systems considerably improved the performance over the provided baseline. Although, the existing gap between the best performing model, i.e., DeepBlueAI \cite{Chunguang2020BERTbased}, and human-level performance shows that more research is required.

\section{Conclusion}
In this paper, we summarized the task of acronym identification and acronym disambiguation at a scientific document understanding workshop (AI@SDU and AD@SDU). For these tasks, we provide two novel datasets that address the limitations of the prior work. Both tasks attracted substantial participants with considerable performance improvement over provided baselines. However, the lower performance of the best performing models compared to the human level performance shows that more research should be conducted on both tasks. 

\bibliography{ref.bib}

\end{document}